\documentclass[runningheads]{llncs}
\usepackage[T1]{fontenc}
\usepackage{graphicx}
\usepackage{booktabs}
\usepackage[misc]{ifsym}
\newcommand{\corr}{(\Letter)}
\usepackage{xcolor}
\usepackage{tabularx}
\newcolumntype{C}{>{\centering\arraybackslash}X}
\newcolumntype{R}{>{\raggedleft\arraybackslash}X}
\newcolumntype{L}{>{\raggedright\arraybackslash}X}
\usepackage{array}
\usepackage{makecell}
\usepackage{float}
\usepackage{subcaption}
\usepackage{url}

\makeatletter
\newcommand\notsotiny{\@setfontsize\notsotiny{6.31415}{7.1828}}
\makeatother

\begin{document}

\title{How to RETIRE Tabular Data in Favor \\of Discrete Digital Signal Representation}

\titlerunning{How to RETIRE Tabular Data in Favor of Discrete Digital Signal \ldots}

\author{
Pawe{\l} Zyblewski\orcidID{0000-0002-4224-6709} \corr \and\\
Szymon Wojciechowski\orcidID{0000-0002-8437-5592}}
\authorrunning{P. Zyblewski and S. Wojciechowski}

\institute{Department of Systems and Computer Networks, 
            Faculty of Information and Communication Technology, 
            Wroc{\l}aw University of Science and Technology,
            Wybrze{\.z}e Wyspia{\'n}skiego 27, 50-370 Wroc{\l}aw, Poland\\ \email{\{pawel.zyblewski;szymon.wojciechowski\}@pwr.edu.pl}}

\maketitle

\begin{abstract}
The successes achieved by deep neural networks in computer vision tasks have led in recent years to the emergence of a new research area dubbed \emph{Multi-Dimensional Encoding} (\textsc{mde}). Methods belonging to this family aim to transform tabular data into a homogeneous form of discrete digital signals (images) to apply convolutional networks to initially unsuitable problems. Despite the successive emerging works, the pool of \emph{multi-dimensional encoding} methods is still low, and the scope of research on existing modality encoding techniques is quite limited. To contribute to this area of research, we propose the \emph{Radar-based Encoding from Tabular to Image REpresentation} (\textsc{retire}), which allows tabular data to be represented as radar graphs, capturing the feature characteristics of each problem instance. \textsc{retire} was compared with a pool of \emph{state-of-the-art} \textsc{mde} algorithms as well as with \emph{XGBoost} in terms of classification accuracy and computational complexity. In addition, an analysis was carried out regarding transferability and explainability to provide more insight into both \textsc{retire} and existing \textsc{mde} techniques. The results obtained, supported by statistical analysis, confirm the superiority of \textsc{retire} over other established \textsc{mde} methods.

\end{abstract}

\section{Introduction}Even though current technological developments are leading to a continuous increase in the amount of data in text or discrete digital signal form, tabular modality is still considered the most popular form of data \cite{shwartz2022tabular}, commonly found in tasks such as medical diagnosis \cite{ulmer2020trust}, recommendation systems \cite{zhang2021neural}, cybersecurity \cite{buczak2015survey} or psychology \cite{urban2021deep}. This information is essential because, despite the increasing proliferation of deep learning methods that excel at computer vision tasks or audio and text analysis, it is tabular data that provides enough of a challenge for it to be called the “last unconquered castle” for deep neural networks \cite{kadra2021well}. This is due to the heterogeneous nature of tabular data, which, unlike text, discrete digital signals or acoustic signals -- which are inherently homogeneous in nature -- can contain different feature types. In brief, the literature currently identifies four significant challenges in analyzing tabular data: \textbf{(i)} Low-Quality Training Data, \textbf{(ii)} Missing or Complex Irregular Spatial Dependencies, (iii) Dependency on Preprocessing, and \textbf{(iv)} Importance of Single Features \cite{borisov2022deep}.

Due to these difficulties, using deep learning for tabular data analysis is currently an important and rapidly growing area of research. At present, a taxonomy of approaches to the application of deep learning for tabular data distinguishes three main paths: \textbf{(i)} Data Transformation Methods, \textbf{(ii)} Specialized Architectures, and \textbf{(iii)} Regularization Models \cite{borisov2022deep}. While the vast majority of work in this area focuses on specialized network architectures or dedicated tabular data regularization mechanisms, this article focuses on data transformation methods, specifically \emph{Multi-Dimensional Encoding} (\textsc{mde}) techniques. While \emph{Single-Dimensional Encoding} methods, such as ordinal or label encoding, are dedicated to transforming categorical variables into real numbers, \textsc{mde} aims to turn entire feature vectors into image representation in order to facilitate employing convolutional neural networks for initially unsuitable problems. The upside of such approaches is the ability to leverage existing architectures without modification, as well as the potential benefit of \emph{transfer learning} \cite{zhuang2020comprehensive}.

As already mentioned, \emph{multi-dimensional encoding} methods, despite offering a highly intriguing way of dealing with tabular data and presenting potentially valuable results, are a definite minority in the pool of scientific articles addressing the application of deep learning in this area. As a result, the number of available \textsc{mde} methods is relatively low, and their potential is still under-researched. In an effort to take the next step towards filling this gap in the literature and expanding the field of \emph{multi-dimensional encoding}, this paper introduces the \emph{Radar-based Encoding from Tabular to Image REpresentation} (\textsc{retire}) method. \textsc{retire} allows individual instances of tabular datasets to be depicted in the form of radar diagrams, capturing the characteristics of their feature values. The proposed method has been compared based on a robust experimental protocol with \textsc{mde} algorithms known from the literature in order to determine which image representation offers the highest generalization ability while employing the \emph{ResNet18} \cite{he2016deep} as the architecture of choice. In addition, the \emph{XGBoost} \cite{chen2016xgboost} algorithm was used as a strong baseline for tabular data classification tasks. In order to gain better insight into the interaction of selected \textsc{mde} methods with convolutional networks, an attempt was made to analyze them in terms of transferability and explainability.

In brief, the main contributions of this article are as follows: 
\begin{itemize}
    \item The introduction of a novel \textit{Radar-based Encoding from Tabular to Image REpresentation} (\textsc{retire}) \emph{multi-dimensional} encoding method.
    \item Extensive experimental evaluation of \textsc{retire} on $22$ benchmark datasets, including selected modality encoding methods and the \emph{XGBoost} algorithm.
    \item Analysis of selected \emph{state-of-the-art} \emph{multi-dimensional encoding} methods and \textsc{retire} in terms of transferability, explainability and computational complexity.
\end{itemize}

\section{Related works}
\label{Section2}
This chapter briefly introduces existing approaches to applying deep neural networks to the task of tabular data classification, with an emphasis on multi-dimensional encoding methods. In addition, to highlight the research area's potential and the scientific community's interest, the \emph{Sentence Space} approach, which is similar to \textsc{mde} but applied to textual modality, is described.

\subsection{Deep Neural Networks for Tabular data}
As already mentioned, the taxonomy for employing deep learning for tabular data distinguishes three groups: \textbf{(i)} Data Transformation Methods, \textbf{(ii)} Specialized Architectures, and \textbf{(iii)} Regularization Models \cite{borisov2022deep}. While the \emph{multi-dimensional encoding} methods are the main focus of this article, which is described in a separate section, the other approaches are briefly introduced below.

Specialized architectures, the largest group of approaches, focus on developing deep neural network structures dedicated to tabular data and considering its heterogeneous nature. The literature here distinguishes two subgroups: \textbf{(i)} Hybrid models and \textbf{(ii)} Transformer-based models. Solutions from the hybrid models group combine canonical machine learning algorithms with neural networks and include both fully and partly differentiable models. The Network-On-Network (\textsc{non}) classification model by Luo et al. \cite{luo2020network} consists of a fieldwise network containing unique deep networks for each problem feature, a cross-field network choosing optimal operations for a given dataset, and an operation fusion network, allowing for nonlinearities by connecting selected operations. In contrast, Ke et al. proposed the \emph{DeepGBM} model, which combines neural networks with the preprocessing capability of gradient-boosted decision trees. \emph{DeepGBM} consists of two networks, one dealing with dense numerical features and the other with sparse categorical features\cite{ke2019deepgbm}. A subgroup of transformer-based approaches is inspired by transformers' successes in text and image data tasks \cite{khan2022transformers}. One of the most popular examples of a model in this category is \emph{TabNet}, introduced by Arik and Pfister \cite{arik2021tabnet}, consisting of multiple sequential hierarchical subnetworks, where each corresponds to a single decision step.

Regularization models are based on the assumption that in the case of heterogeneous tabular data, the flexibility of neural networks can be an obstacle, and strong regularization of parameters is needed. Kadra et al. \cite{kadra2021well} have shown that a multilayer perceptron using a set of 13 different regularization methods can outperform \emph{state-of-the-art} tabular data classification methods at the cost of increased time.

\emph{Single-Dimensional Encoding}, which belongs to the Data Transformation methods group, focuses on dealing with categorical features by encoding them into a form suitable for deep neural networks \cite{hancock2020survey}. \emph{Ordinal encoding}, \emph{label encoding}, and \emph{one-hot encoding} are among the most popular approaches. We can also distinguish \emph{leave-one-out encoding}, which replaces each category with the target variable's average value, and \emph{hash-based encoding}, which transforms each category using a hash function to a fixed-size value.

\subsection{Multi-dimensional Encoding}
Numerous scientific articles attest to the effectiveness of convolutional networks in both unimodal and multimodal scenarios for natural language analysis \cite{gimenez2020semantic} and image, video, and audio classification (e.g., in spectrogram form) \cite{satt2017efficient}. Additionally, deep networks facilitate \emph{transfer learning}, which enables models to apply previously learned information to the current challenge \cite{zhuang2020comprehensive}. As a result of these achievements, a new field of study known as \emph{Multi-Dimensional Encoding} \cite{borisov2022deep} was created that focuses on converting tabular data into a more uniform form of digital discrete signals.

\begin{figure}[!htb]
\centering
 \includegraphics[width=.8\textwidth]{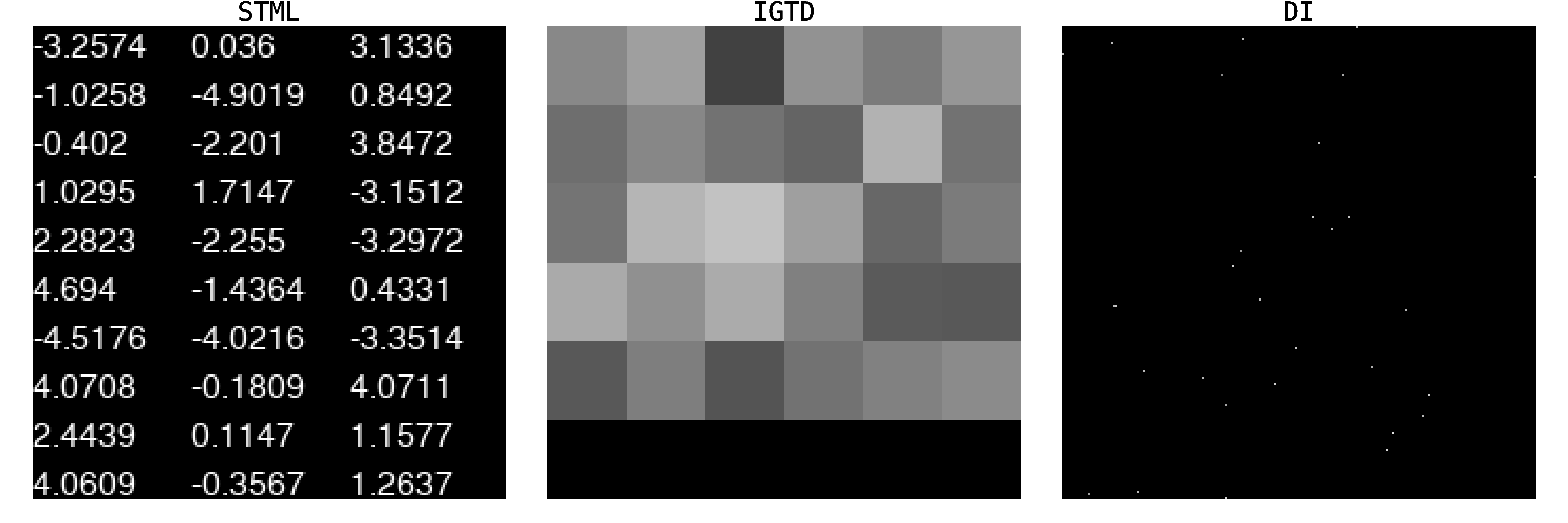}
\caption{An example of encoding a single sample of a synthetically generated tabular problem with 30 features, into a two-dimensional discrete digital signal using \textsc{stml}, \textsc{igtd} and \textsc{di} techniques.}
\label{fig:mde}
\end{figure}

This idea is best shown by \emph{Super Tabular Machine Learning} (\textsc{stml}) approach, proposed by Sun et al. \cite{sun2019supertml}, which projects the feature values of a given problem instance onto the image. The \emph{Image Generator for Tabular Data} (\textsc{igtd}) \cite{zhu2021converting} created by Zhu et al. offers an alternative approach. By minimizing the discrepancy between the ranking of distances among features and the ranking of distances between their corresponding pixels in the image, the algorithm looks for an optimal mapping. Damri et al. proposed \emph{Feature Clustering-Visualization} (\emph{FC-Viz}) \cite{damri2023towards} approach, where each instance of tabular data is converted into a 2D pixel-based representation, where pixels representing strongly correlated and interacting features are located in close proximity of each other. \emph{SuperTML-Clustering}, as proposed by Zhang and Ding, embeds the indices of clustered continuous feature values onto an image \cite{zhang2023supertml}. \emph{DeepInsight} (\textsc{di}) by Sharma et al. \cite{sharma2019deepinsight} allows for the collective utilization of neighboring elements by arranging different elements or features farther away and similar elements or features closer together. The exemplary results of employing selected multi-dimensional encoding methods to tabular data is shown in Fig. \ref{fig:mde}. It is also worth noting that recently, there have been first works successfully employing \textsc{mde} in data stream classification tasks while demonstrating relatively low computation time \cite{zyblewski2024employingSTML}.

Although \emph{multi-dimensional encoding} methods were developed to convert heterogeneous tabular data into homogeneous discrete digital signals, similar approaches have also been successfully applied to inherently homogeneous text modality with remarkable results. Such techniques are usually based on \emph{Sentence Space}, which is a reference to the method presented by Kim \cite{kim2014convolutional}, in which text data is converted into an image with embeddings of individual words in each row. While these methods are not the subject of this article, and we will not describe them in depth, they undoubtedly provide evidence of the scientific community's interest in \textsc{mde} methods and their derivatives.

\section{Radar-based Encoding from Tabular to Image REpresentation (RETIRE)}
The \textsc{retire} algorithm is inspired by radar charts -- a form of graphical representation of an instance's attributes. It can be often seen in cases where one need to compare available options with each other based on their properties. A radar chart is constructed by drawing a polygon over coordinates in a polar coordinate system. Such a figure has $N$ vertices (same as object attributes), which
coordinates ($r$ - radius, $\phi$ - angle) can be determined as:

$$
r_n = f_n, \quad \phi_n = (n - 1) \times \frac{2\pi}{N}
$$

Where $f_n$ is the value of an $n$-th attribute. Obviously, for the chart to be readable, values have to be relatable to each other. To achieve that, the scaling function $S$ is often used so that $\{ S(f_n) \quad | \quad l <= S(f_n) <= u \}$ where $l$ and $u$ is a set parameter (usually $l = 0, u = 1$).

\begin{figure}[!htb]
\centering
\begin{subfigure}[b]{0.8\textwidth}
\centering
    \includegraphics[width=0.8\textwidth]{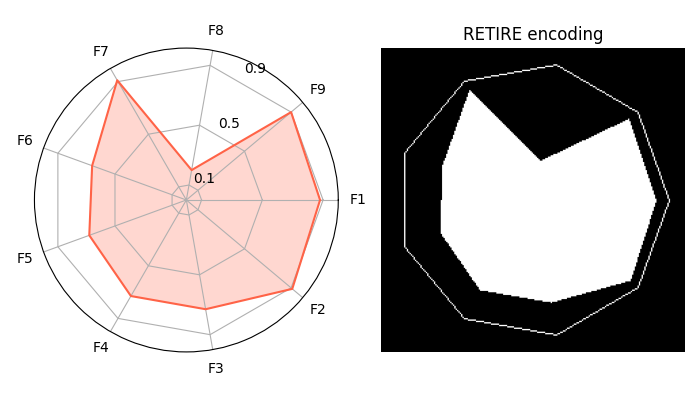}
    \caption{Example of radar chart (left) and \textsc{retire} encoding (right).}
    \label{fig:retire_example}
\end{subfigure}%

\begin{subfigure}{0.7\textwidth}
\centering
    \includegraphics[width=\textwidth]{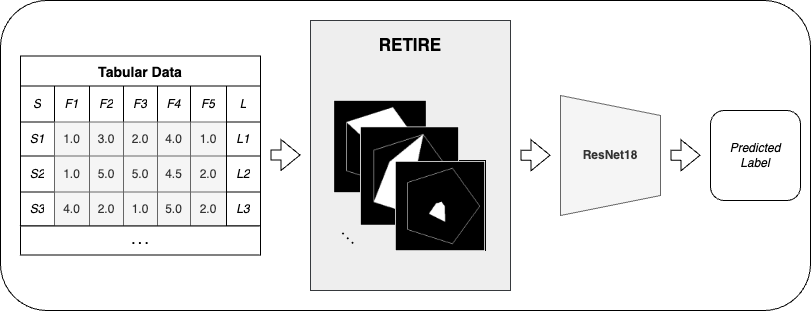}
    \caption{Pipeline for classification system.}
    \label{fig:convdiag}
\end{subfigure}

\caption{Graphical abstract for the proposed method.}
\label{fig:abstract}
\end{figure}

The radar charts often include the axis grid and labels. The figure can also introduce colors if used to compare multiple samples. The goal of the encoding algorithm, however, is to create a representation in the shortest possible time, so the \textsc{retire} algorithm extracts only the binarized shape of the sample and a border indicating the upper limit of the value. An example of the original graph and the corresponding representation is presented in Fig. \ref{fig:retire_example}.

Of course, there can be many modifications to the proposed method, which are the parameters of the algorithm. The very important one is assumed scaling. In this work, $S$ will be function applied to each features as:

$$
S(x_f) = \frac{(x_f - min(X_f))}{(max(X_f) - min(X_f))} \times (u - l) + l
$$

Where $X$ is the full set of observations and $x$ is a single sample. As can be seen, the scaling requires obtaining minimum and maximum values of $X$. Those have to be determined during the learning process -- which indicates that \textsc{retire} requires training before being applied to the dataset.

In addition, $u$ and $l$ are also important scaling parameters. Since the first part of the equation will be in the range of $<0, 1>$, setting $l >0$ and $u < 1$ allows scaling to provide “space” for observations. It is important because when \textsc{retire} is used for test data, values from the test set might be outside the observed min-max range. Finally, an important element is the retained figure outline -- it provides a reference point for the maximum observed value. The full pipeline of the proposed method is shown in Fig. \ref{fig:convdiag}.

The image representations obtained utilizing encoding are then used by \textsc{cnn} -- in this case, \emph{ResNet18}. It is worth mentioning that although the image created by \textsc{retire} is binarized, \textsc{cnn} uses all color channels as the input. Therefore, the resulting binary image is multiplied to obtain 3 identical color channels.

\section{Experimental Evaluation}
This section describes in detail the experimental evaluation process conducted to analyze the properties of the proposed \textsc{retire} method. The experiments were designed to answer the following research questions:
\begin{itemize}
    \item \textbf{RQ1} Does the discrete digital signal representation obtained using \textsc{retire} benefits from ImageNet knowledge transfer in terms on \emph{TransRate} metric?
    \item \textbf{RQ2} In the case of the \emph{ResNet18} architecture, does the use of an image representation obtained using \textsc{retire} offers a statistically significantly better balanced accuracy score compared to reference \textsc{mde} methods?
    \item \textbf{RQ3} Can the \textsc{retire} image representation be interpreted by a human?
    \item \textbf{RQ4} What is the computational complexity of the \textsc{retire} encoding and how it compares with reference methods?
\end{itemize}

\subsection{Set-up}
\textbf{Data} The experiments were conducted using 22 benchmark datasets originating from the \textsc{keel} repository \cite{derrac2015keel}. The datasets can be considered balanced - most have an \emph{Imbalance Ratio} (\textsc{ir}) of less than 1.5, and only in one case the \textsc{ir} exceeds 2. The precise dataset characteristics are shown in Tab. \ref{tab:datasets}.

\begin{table}[!htb]
    \centering
    \notsotiny
    \caption{Datasets characteristics.}
    \begin{tabularx}{0.99\textwidth}{LCC|LCC}
    \toprule
    Dataset & \#Instances & \#Features & Dataset & \#Instances & \#Features \\
    \midrule
    australian & 690 & 14 & mammographic & 830 & 5 \\
    banknote & 1372 & 4 & monk-2 & 432 & 6 \\
    breastcan & 683 & 9 & monkone & 556 & 6 \\
    breastcancoimbra & 116 & 9 & phoneme & 5404 & 5 \\
    bupa & 345 & 6 & pima & 768 & 8 \\
    cryotherapy & 90 & 6 & ring & 7400 & 20 \\
    german & 1000 & 24 & sonar & 208 & 60 \\
    haberman & 306 & 3 & spambase & 4601 & 57 \\
    heart & 270 & 13 & titanic & 2201 & 3 \\
    ionosphere & 351 & 34 & twonorm & 7400 & 20 \\
    liver & 345 & 6 & wisconsin & 699 & 9 \\
    \bottomrule
    \end{tabularx}
    \label{tab:datasets}
\end{table}

\textbf{Experimental protocol}
In order to guarantee a robust experimental protocol, the experiments were carried out using 5-times repeated 2-fold stratified cross-validation. The results were supported by the \emph{Combined 5x2 CV F-test} \cite{alpaydm1999combined} and the \emph{Wilcoxon signed-rank test} \cite{stapor2021design} with $\alpha=0.05$ (the higher the rank, the better). \emph{The Combined 5x2 CV F-test} was used to examine the statistical correlations between methods within particular datasets, whilst the \emph{Wilcoxon signed-rank test} allowed for a global comparison of the investigated approaches. Although the datasets used have a relatively low \textsc{ir} and can be considered balanced, the classification performance was evaluated based on the \emph{balanced accuracy score} (\textsc{bac}).

\textbf{Reference methods} In the course of the experiments, \textsc{retire} was compared with three state-of-the-art \textsc{mde} approaches described in Section \ref{Section2}: \textbf{(i)} \emph{SuperTML}, \textbf{(ii)} \textsc{igtd}, and \textbf{(iii)} \emph{DeepInsight}. They were chosen because of the promising results obtained in previous studies and the access to a Python implementation offered by their authors (\textsc{igtd} and \textsc{di}). Despite the lack of an official implementation, we implemented \textsc{stml} due to its simplicity. \textsc{igtd} and \textsc{di} were used with default parameters. The size of the image representation for \textsc{stml}, \textsc{di}, and \textsc{retire} multi-dimensional encoding techniques was set to 224x224 pixels. The size of images created using \textsc{igtd}, due to the characteristics of the method, depended on the number of features in each dataset. For each method the resulting representation is multiplied in order to achieve 3 identical color channels.

Regardless of the \textsc{mde} approach used, the \emph{ResNet18} architecture was chosen as a convolutional network due to its vast popularity and relatively small size. Training lasted for 20 epochs and the batch size was set to $8$. An \textsc{sgd} optimizer with a learning rate of $0.001$ and momentum of $0.9$ was used. Cross-entropy loss was chosen as the loss function.

In addition, the \emph{XGBoost} \cite{chen2016xgboost} algorithm with 100 estimators was used as the baseline. The other parameters remained default. This comparison was added to improve the interpretability of the results obtained through experiments, but it should be noted that \emph{XGBoost} is a very strong benchmark and winning against it is not the primary goal of this work.

\textbf{Reproducibility}
All experiments presented in this paper were carried out in \emph{Python} using the \emph{scikit-learn} \cite{scikit-learn}, \emph{PyTorch} \cite{paszke2017automatic}, and \emph{XGBoost} \cite{chen2016xgboost} libraries. The results obtained can be fully replicated using the code located on the \emph{GitHub} repository\footnote{\url{https://github.com/w4k2/mde-retire}}. The computing platform used in all experiments was Mac Studio with Apple M1 Ultra with 20‑core CPU, 64‑core GPU, 32‑core Neural Engine system, 128 GB RAM.

\subsection{Experimental scenarios}
\noindent\textbf{Experiment 1 -- Comparative experimental evaluation}
The purpose of Experiment 1 is to investigate which \textsc{mde} method achieves statistically significantly the best balanced accuracy score for the 22 selected datasets and how it compares to the results obtained by the \emph{XGBoost} algorithm. This is the main experiment of this paper, and it will allow us to investigate the quality of the image representations offered by each \textsc{mde} method, thus answering \textbf{RQ2}.

Additionally, in order to verify the usefulness of \emph{transfer learning} in conjunction with the studied representations obtained using \textsc{mde}, all four encoding methods are evaluated in terms of \emph{TransRate} metrics \cite{huang2022frustratingly} and \textsc{bac}. The \emph{ResNet18} architecture offers the possibility of using pre-trained weights, but many scientific articles utilize this option without considering the possibility of negative transfer \cite{wang2019characterizing}. Since we are concerned with establishing the network's inherent properties, \emph{ResNet18} is not trained or finetuned in any way during this experiment. We only evaluate the initial compatibility of knowledge transfer with individual image representations. The results obtained will make it possible to answer \textbf{RQ1}.

\noindent\textbf{Experiment 2 -- Explainability}
As part of Experiment 2, the performance of the \textsc{mde} algorithms offering representations potentially understandable to humans, namely \textsc{stml} and \textsc{retire}, will be subjected to explainability methods derived from the \textsc{shap} (\emph{SHapley Additive exPlanations}) library \cite{shap1_NIPS2017_7062,shap2_lundberg2020local2global}. The results of this analysis will be contrasted with the explained output of the \emph{XGBoost} algorithm in order to investigate whether humans can interpret the images resulting from the \textsc{mde} and whether the information relevant to the \emph{ResNet18} convolutional network overlaps with that which most influences the decision of the algorithm based solely on tabular data. Based on the observations made, it will be possible to find an answer to \textbf{RQ3}.

\noindent\textbf{Experiment 3 -- Computational complexity analysis} 
The purpose of the experiment is to study and analyze the computational complexity of selected methods. The main factor subject to this study is to demonstrate the relation of inference time to the dimensionality of the problem. For this experiment, the synthetically generated datasets of a variable number of features will be used. Other parameters will be set as follows: \textsc{madelon} generator, 100 samples, informative-only features, and random generation seed for each instance. Other parameters remain the default for each instance. The test will include measurement of encoding time and \emph{ResNet18} model inference. To stabilize the results, the measurements will be repeated 100 times. The results obtained in this externality will make it possible to find the answer to \textbf{RQ4}.

\subsection{Experiment 1 -- Comparative experimental evaluation}
\begin{figure}[!htb]
    \centering
    \includegraphics[width=\textwidth]{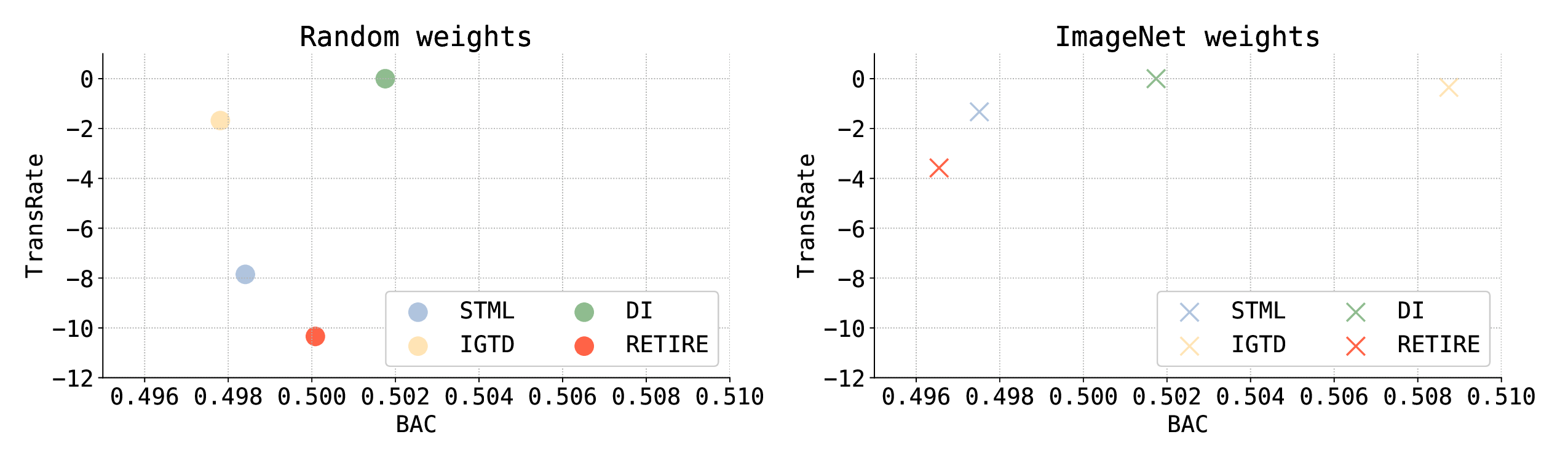}
    \caption{Results of transferability estimation using the \emph{TransRate} metric. Both \textsc{bac} and \emph{TransRate} values were averaged for all 22 datasets.}
    \label{fig:transfer}
\end{figure}

Before the actual comparison experiment, a brief study was conducted to determine whether \textsc{mde} representations could benefit from knowledge transfer derived from the \emph{ImageNet} dataset. The results obtained in terms of \textsc{bac} and \emph{TransRate}, which measures the transferability as the mutual information between features of target examples extracted by a pre-trained model and their labels \cite{huang2022frustratingly}, are shown in Fig. \ref{fig:transfer}.

It should be noted that although \emph{TransRate} has a relatively linear relationship to generalization ability, in this case, we are dealing with a random classifier, and \textsc{bac} values oscillating around $50\%$ regardless of the use of \emph{transfer learning} are expected. This is because, despite the use of \emph{transfer learning}, the images in ImageNet deviate too much from the representations offered by \textsc{mde}.

At the same time, interesting conclusions can be drawn from the obtained \emph{TransRate} values, which clearly divide the four analyzed methods into two groups. \textsc{igtd} and \textsc{di} theoretically show better compatibility with \emph{ResNet18}, but at the same time, the use of pretreated weights has virtually no effect on them. This is probably due to the abstract nature of the representations they offer, which cannot be compared to the images found in \emph{ImageNet}.

On the other hand, \textsc{stml} and \textsc{retire}, using a less abstract two-dimensional word embedding or geometric shape as the basis for encoding, despite theoretically lower compatibility with \emph{ResNet18}, show significant improvement after using pretreated weights. \textsc{retire}'s lower \emph{TransRate} than \textsc{stml} may be explained by its use of encoding based on a single geometric shape, which inherently has a lower level of complexity than directly transcribing feature values onto the image. Without fine-tuning, such representation may carry slightly less information -- especially for a large number of features.

Based on these results, we can answer \textbf{RQ1} and conclude that the \textsc{retire} representation allows us to benefit from positive knowledge transfer from the \emph{ImageNet} dataset. At the same time, due to the lack of negative transfer in the case of all tested \textsc{mde} methods, the pre-trained \emph{ResNet18} is used in further research.

\begin{figure}[!htb]
    \centering
    \includegraphics[width=\textwidth]{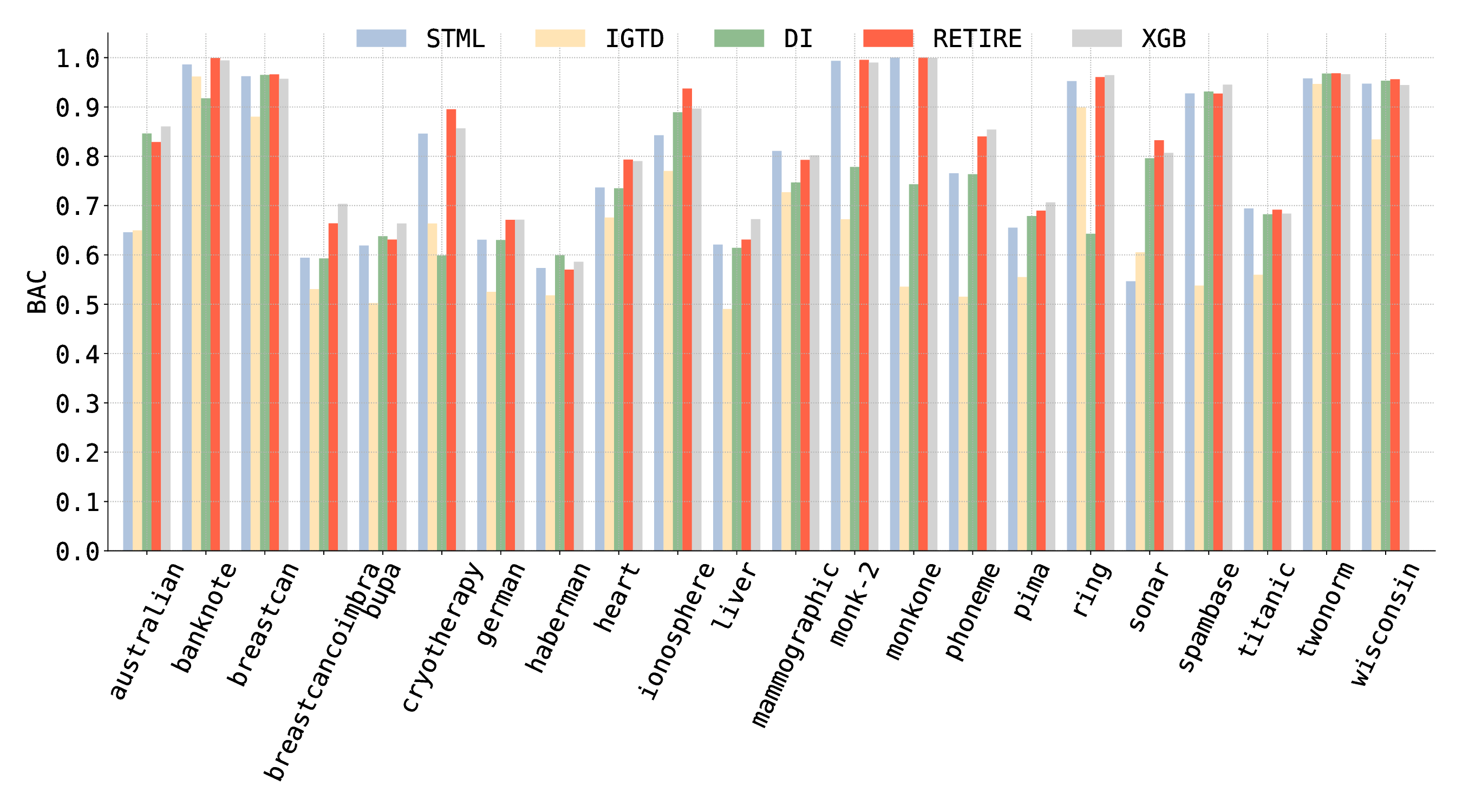}
    \caption{The results of the experimental evaluation conducted in terms of \textsc{bac} for each of the 22 datasets.}
    \label{fig:comparison}
\end{figure}

The results of the experiment comparing \textsc{retire} with the reference \textsc{mde} methods and \emph{XGBoost} can be seen in Fig. \ref{fig:comparison} and in Tab. \Ref{tab:comparison}. The first thing that catches the eye is the relatively low results achieved by \textsc{igtd}. This is because the images produced by this method are relatively small, and their size depends directly on the number of features. Consequently, the \emph{ResNet18}, adapted initially for 224x224 pixels images, is not the optimal solution. However, since the study aimed to evaluate the role of individual image representations in learning one selected architecture, we decided to utilize this method in its original form without any modifications.

\begin{table}[!htb]
    \renewcommand{\arraystretch}{.9}
    \centering
    \notsotiny
    \caption{Results of statistical analysis. The first row for each dataset shows the averaged \textsc{bac}. The indices (by column) of the algorithms from which the given model is statistically significantly better based on the \emph{$5x2$ CV F-test} ($\alpha=0.05$) are given below. The last two rows present the Wilcoxon signed-rank test. The highest \textsc{bac} or rank values are \textbf{bold} or \textcolor{red}{\textbf{highlighted in red}}. \textbf{Bold} indicates the highest value among \textsc{mde} methods, and \textcolor{red}{\textbf{red}} indicates the highest value overall (including \emph{XGBoost}).}

    \begin{tabularx}{\textwidth}{lCCCC|C}
    \toprule
                  & STML$^1$   & IGTD$^2$   & DI$^3$    & RETIRE$^4$   & XGB$^5$     \\
    \midrule
 australian       & 0.646  & 0.650  & \textbf{0.846} & 0.829    & \textcolor{red}{\textbf{0.861}}   \\
                  & ---    & ---    & 1, 2  & 1, 2     & 1, 2, 4 \\
 banknote         & 0.986  & 0.962  & 0.918 & \textcolor{red}{\textbf{0.999}}    & 0.994   \\
                  & 2      & ---    & ---   & 1, 2, 5  & 2       \\
 breastcan        & 0.962  & 0.880  & 0.965 & \textcolor{red}{\textbf{0.966}}    & 0.957   \\
                  & ---    & ---    & ---   & ---      & ---     \\
 breastcancoimbra & 0.594  & 0.531  & 0.593 & \textbf{0.664}    & \textcolor{red}{\textbf{0.703}}   \\
                  & ---    & ---    & ---   & 2        & 1, 2, 3 \\
 bupa             & 0.619  & 0.502  & \textbf{0.638} & 0.631    & \textcolor{red}{\textbf{0.664}}   \\
                  & 2      & ---    & 2     & 2        & 1, 2    \\
 cryotherapy      & 0.846  & 0.664  & 0.599 & \textcolor{red}{\textbf{0.895}}    & 0.857   \\
                  & 2, 3   & ---    & ---   & 2, 3     & 2, 3    \\
 german           & 0.631  & 0.525  & 0.630 & \textcolor{red}{\textbf{0.671}}    & \textcolor{red}{\textbf{0.671}}   \\
                  & 2, 3   & ---    & 2     & 2        & 2, 3    \\
 haberman         & 0.574  & 0.518  & \textcolor{red}{\textbf{0.599}} & 0.570    & 0.586   \\
                  & ---    & ---    & ---   & ---      & 2       \\
 heart            & 0.737  & 0.676  & 0.735 & \textcolor{red}{\textbf{0.793}}    & 0.790   \\
                  & ---    & ---    & ---   & 1, 2     & 1, 2    \\
 ionosphere       & 0.843  & 0.770  & 0.889 & \textcolor{red}{\textbf{0.937}}    & 0.897   \\
                  & 2      & ---    & 1, 2  & 1, 2, 3  & 1, 2    \\
 liver            & 0.621  & 0.490  & 0.614 & \textbf{0.631}    & \textcolor{red}{\textbf{0.672}}   \\
                  & 2      & ---    & 2     & 2        & 2       \\
 mammographic     & 0.811  & 0.727  & 0.747 & \textbf{0.793}    & \textcolor{red}{\textbf{0.802}}   \\
                  & ---    & ---    & ---   & ---      & ---     \\
 monk-2           & 0.993  & 0.672  & 0.778 & \textcolor{red}{\textbf{0.995}}    & 0.990   \\
                  & 2, 3   & ---    & ---   & 2, 3     & 2, 3    \\
 monkone          & \textcolor{red}{\textbf{1.000}}  & 0.536  & 0.743 & \textcolor{red}{\textbf{1.000}}    & 0.999   \\
                  & 2, 3   & ---    & ---   & 2, 3     & 2, 3    \\
 phoneme          & 0.766  & 0.515  & 0.764 & \textbf{0.840 }   & \textcolor{red}{\textbf{0.854}}   \\
                  & 2      & ---    & 2     & 1, 2, 3  & 1, 2, 3 \\
 pima             & 0.655  & 0.555  & 0.679 & \textbf{0.690}    & \textcolor{red}{\textbf{0.706}}   \\
                  & ---    & ---    & 2     & 2, 3     & 1, 2    \\
 ring             & 0.952  & 0.899  & 0.643 & \textbf{0.961}    & \textcolor{red}{\textbf{0.964}}   \\
                  & 2, 3   & 3      & ---   & 3        & 2, 3    \\
 sonar            & 0.546  & 0.605  & 0.796 & \textcolor{red}{\textbf{0.832}}    & 0.807   \\
                  & ---    & ---    & 1, 2  & 1, 2     & 1, 2    \\
 spambase         & 0.927  & 0.538  & \textbf{0.931} & 0.927    & \textcolor{red}{\textbf{0.945}}   \\
                  & 2      & ---    & 2     & 2        & 2, 3, 4 \\
 titanic          & \textcolor{red}{\textbf{0.694}}  & 0.560  & 0.682 & 0.692    & 0.684   \\
                  & 2, 5   & ---    & 2     & ---      & ---     \\
 twonorm          & 0.958  & 0.947  & \textcolor{red}{\textbf{0.968}} & \textcolor{red}{\textbf{0.968}}    & 0.966   \\
                  & 2      & ---    & 2     & 2        & 2       \\
 wisconsin        & 0.947  & 0.834  & 0.953 & \textcolor{red}{\textbf{0.956}}    & 0.945   \\
                  & 2      & ---    & 2     & 2        & 2       \\
    \midrule
    Mean rank & 2.932 & 1.227 & 2.682 & \textcolor{red}{\textbf{4.114}}   & 4.045   \\
      & 2     & ---   & 2     & 1, 2, 3 & 1, 2, 3 \\
    \bottomrule
    \end{tabularx}

    \label{tab:comparison}
\end{table}

Among the other \textsc{mde} methods, \textsc{retire} shows the highest generalization ability by far. For 17 of the 22 datasets, it achieved the highest \textsc{bac} value. For 6 datasets, it proved statistically significantly better than \textsc{stml}, and for 7 datasets - than \textsc{di}. At the same time, none of the reference \textsc{mde} methods ever obtained a statistically significantly better balanced accuracy score than \textsc{retire}.

It is somewhat of a surprise that \textsc{retire} for 11 datasets obtained a higher average \textsc{bac} than \emph{XGBoost}. The difference was statistically significantly better in only one case, but these minor differences translated into \textsc{retire} obtaining a higher average rank value than \emph{XGBoost}. At the same time, \textsc{retire} is the only one of the \textsc{mde} methods studied that, according to the \emph{Wilcoxon signed-rank} test, globally scored statistically comparable to \emph{XGBoost}. Also, like \emph{XGBoost}, \textsc{retire} is globally statistically significantly better than the other reference \textsc{mde} methods, thus responding to \textbf{RQ2}.

\subsection{Experiment 2 -- Explainability}
\begin{figure}[!htb]
\centering
\begin{subfigure}{0.42\textwidth}
\centering
     \includegraphics[width=.99\textwidth]{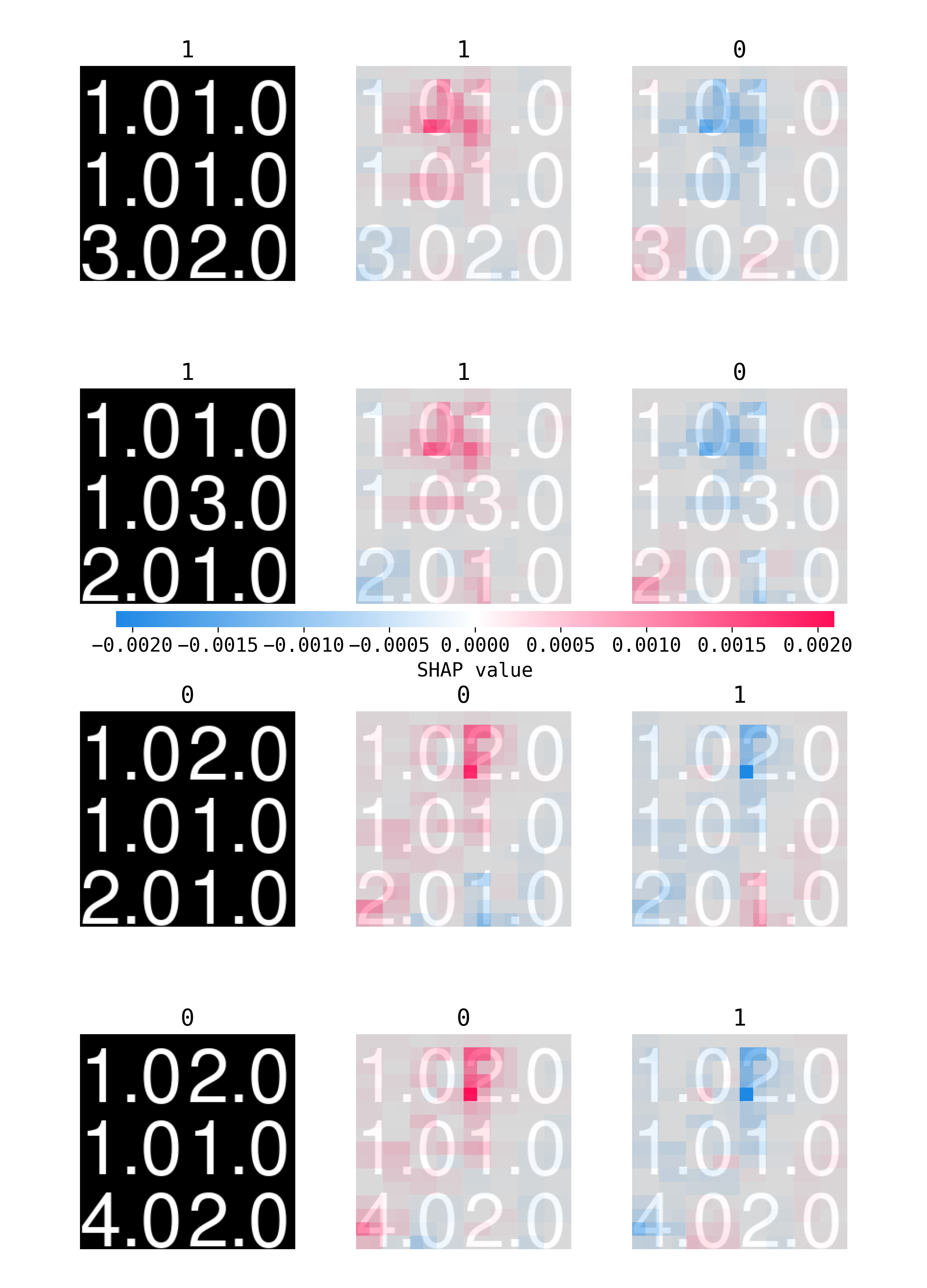}
    \caption{STML}
    \label{fig:one}
\end{subfigure}%
\begin{subfigure}{0.42\textwidth}
\centering
     \includegraphics[width=.99\textwidth]{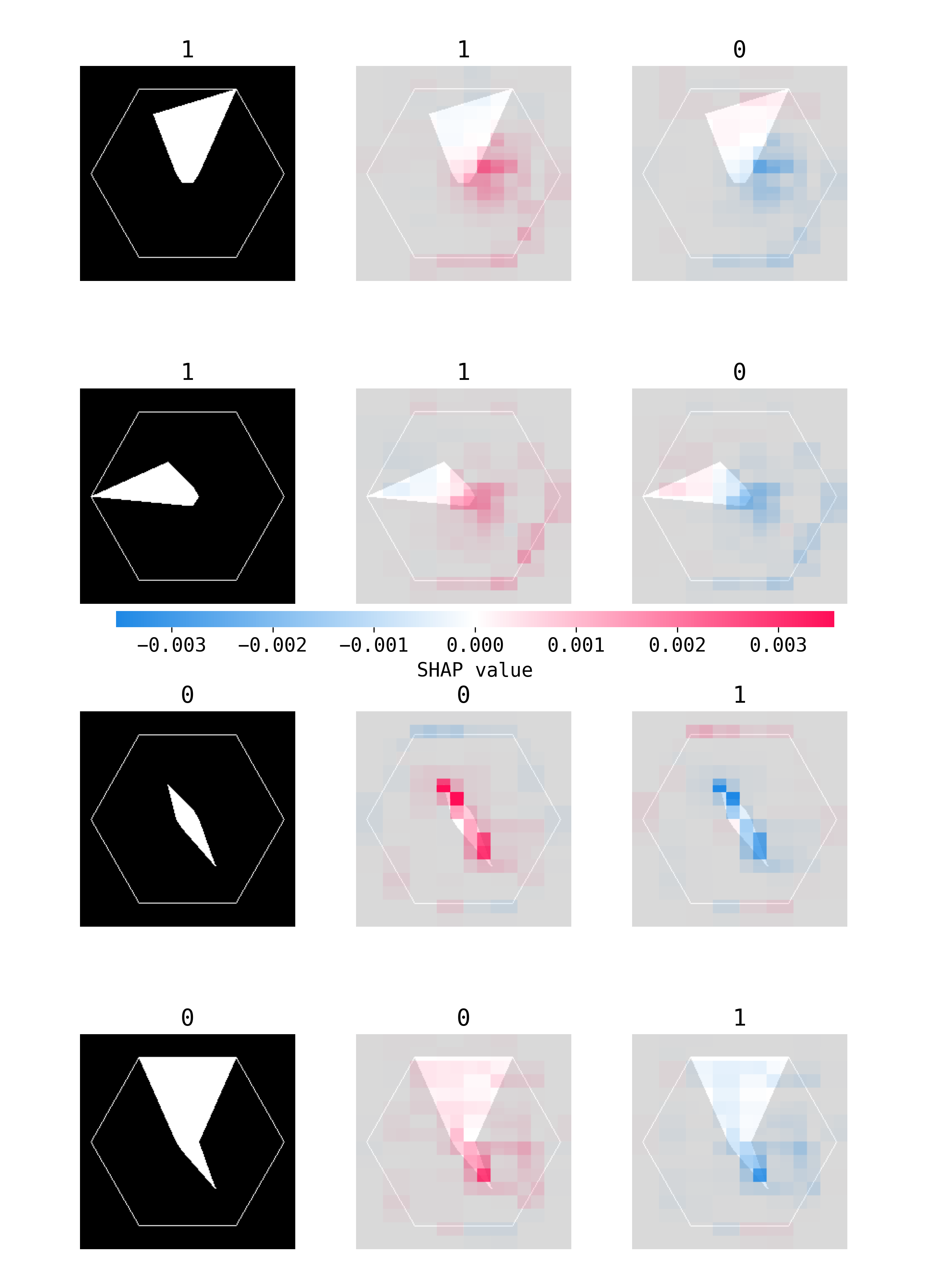}
    \caption{RETIRE}
    \label{fig:two}
\end{subfigure}%

\begin{subfigure}{.9\textwidth}
\centering
     \includegraphics[width=.44\textwidth]{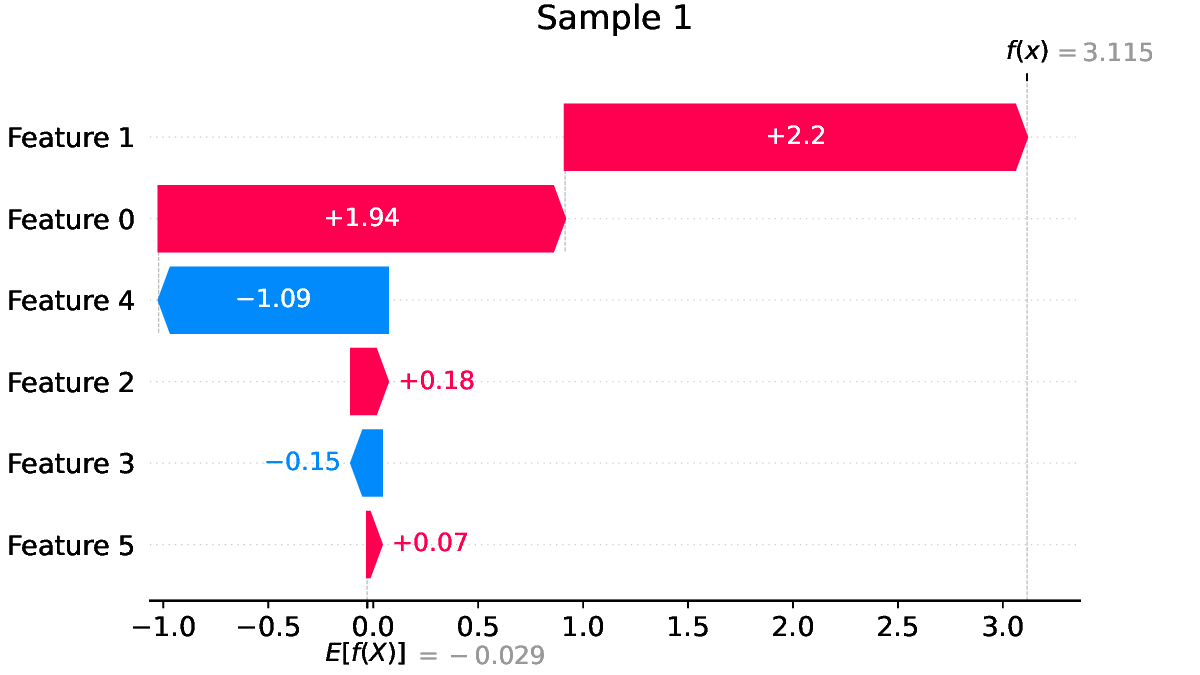}
     \includegraphics[width=.44\textwidth]{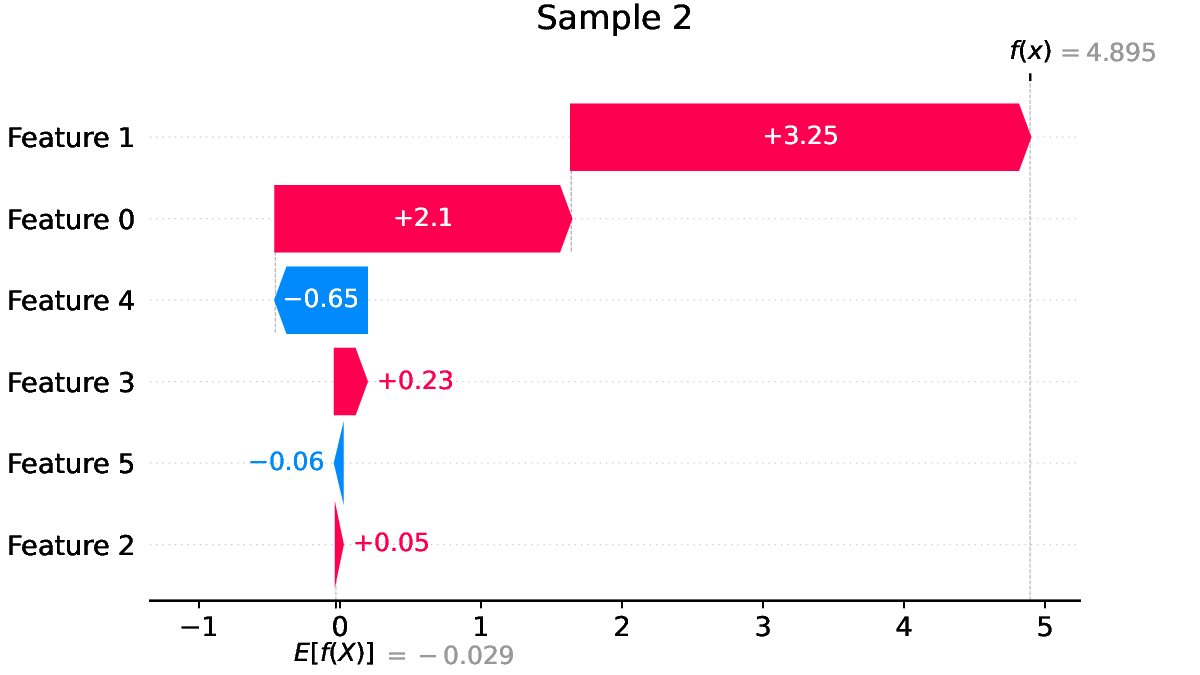}
     \includegraphics[width=.44\textwidth]{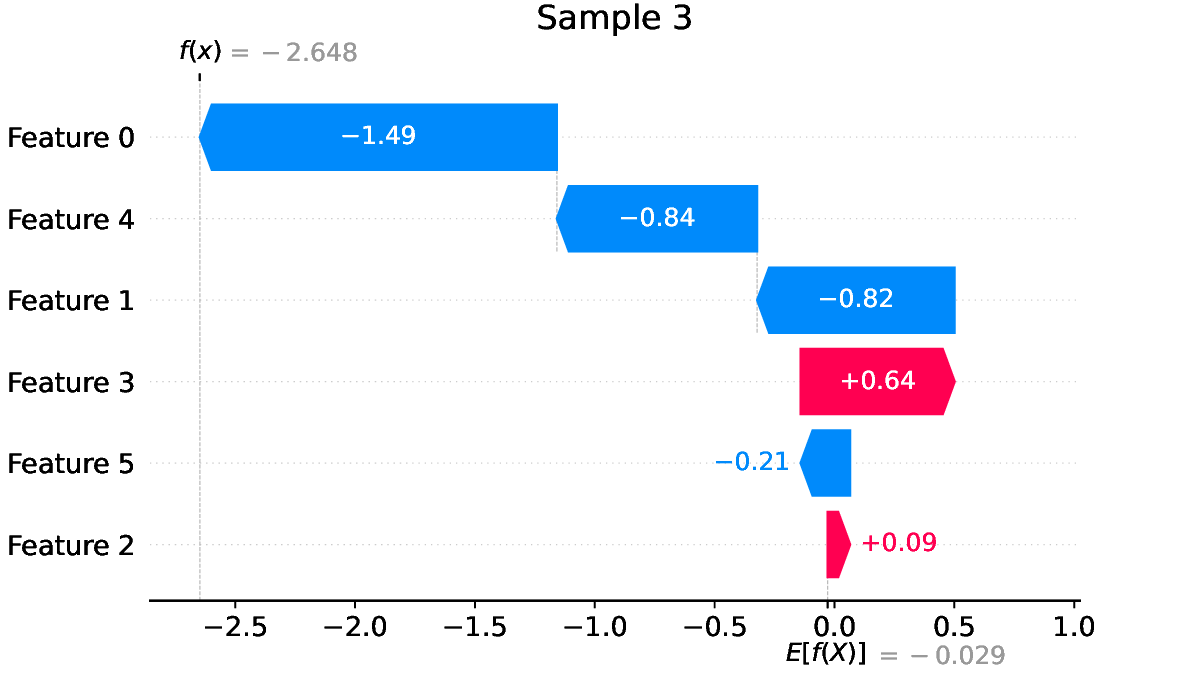}
     \includegraphics[width=.44\textwidth]{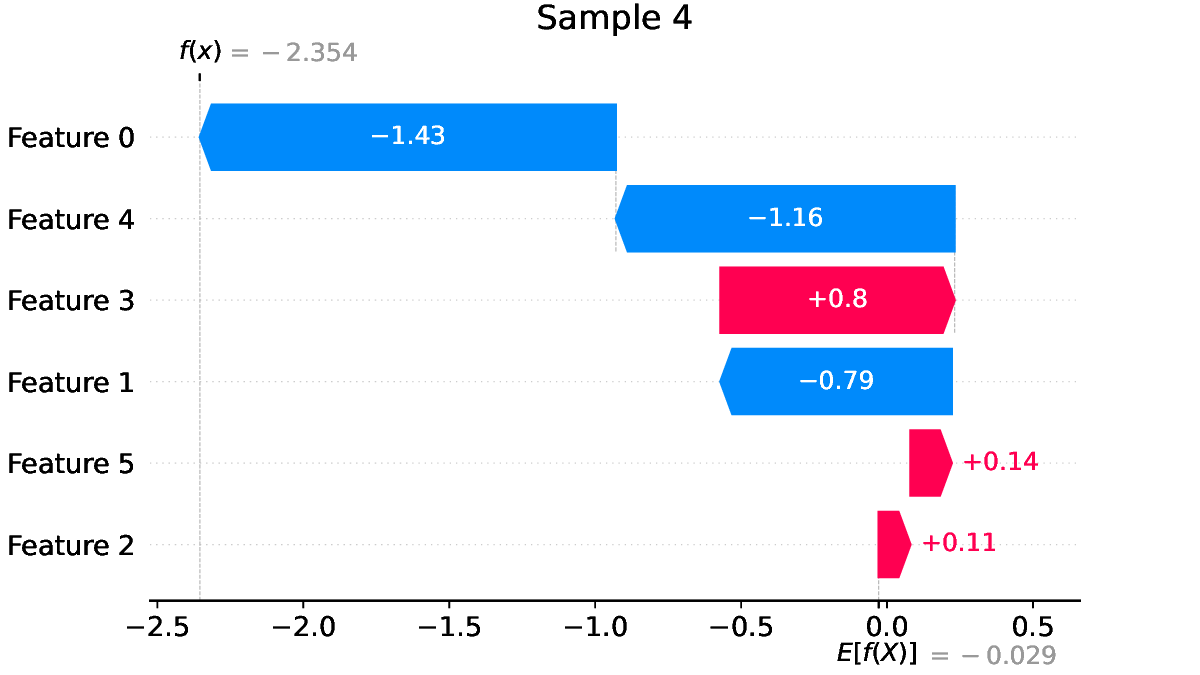}
    \caption{XGBoost}
    \label{fig:three}
\end{subfigure}%

\caption{Explainability of selected data transformation methods applied to the first data fold of the Monkone dataset. The sample numbers in subfigure (c) correspond to the subsequent rows in subfigures (a) and (b). In subfigures (a) and (b), the consecutive columns from the left represent: (i) the original image and its class, (ii) the classifier's decision and the explanation behind it, and (iii) the decision with less support value and its justification.}
\label{fig:xai}
\end{figure}

Fig. \ref{fig:xai} presents the results of the explainability analysis for the \textsc{retire}, \textsc{stml}, and \emph{XGBoost} algorithms in the form of a \textsc{shap} image plot and Waterfall plot. Since the purpose of the analysis was to verify that the features relevant to \emph{ResNet18} in the image representations overlapped with those that determined the \emph{XGBoost} algorithm's decision, the 4 samples analyzed were taken from the \emph{monkone} dataset, for which all three approaches achieved near perfect \textsc{bac}.

In the case of \textsc{stml}, we can see that regardless of the class, the first row of the image containing \textit{Feature 0} and \textit{Feature 1} values strongly influences the decision. In the case of \textit{Sample 3} and \textit{Sample 4}, \textit{Feature 4}, located in the lower-left corner of the representation, also plays a significant role in making the correct decision. We can see similar correlations in the case of \textsc{retire}, where the first two features (starting from the right and going clockwise) seem to be the most significant. In addition, in the case of \textit{Class 0}, the area belonging to \textit{Feature 4} also plays an important role in the decision-making process. The most significant observation here is that the components of the \textsc{stml} and \textsc{retire} images that \emph{ResNet18} found to be most relevant overlap significantly with the features that have the greatest impact on the decision of the \emph{XGBoost} algorithm, based solely on tabular data. This confirms that despite minor differences due to the characteristics of the individual approaches and differing representations, convolutional networks employing \textsc{mde} methods (including the proposed \textsc{retire}) make their decisions based on the values of the problem's features in a manner similar to algorithms operating on tabular data. This makes the proposed representation interpretable to humans, thus answering \textbf{RQ3}.

\subsection{Experiment 3 -- Computational complexity analysis}
\begin{figure}[!htb]
    \centering
    \includegraphics[width=.9\textwidth]{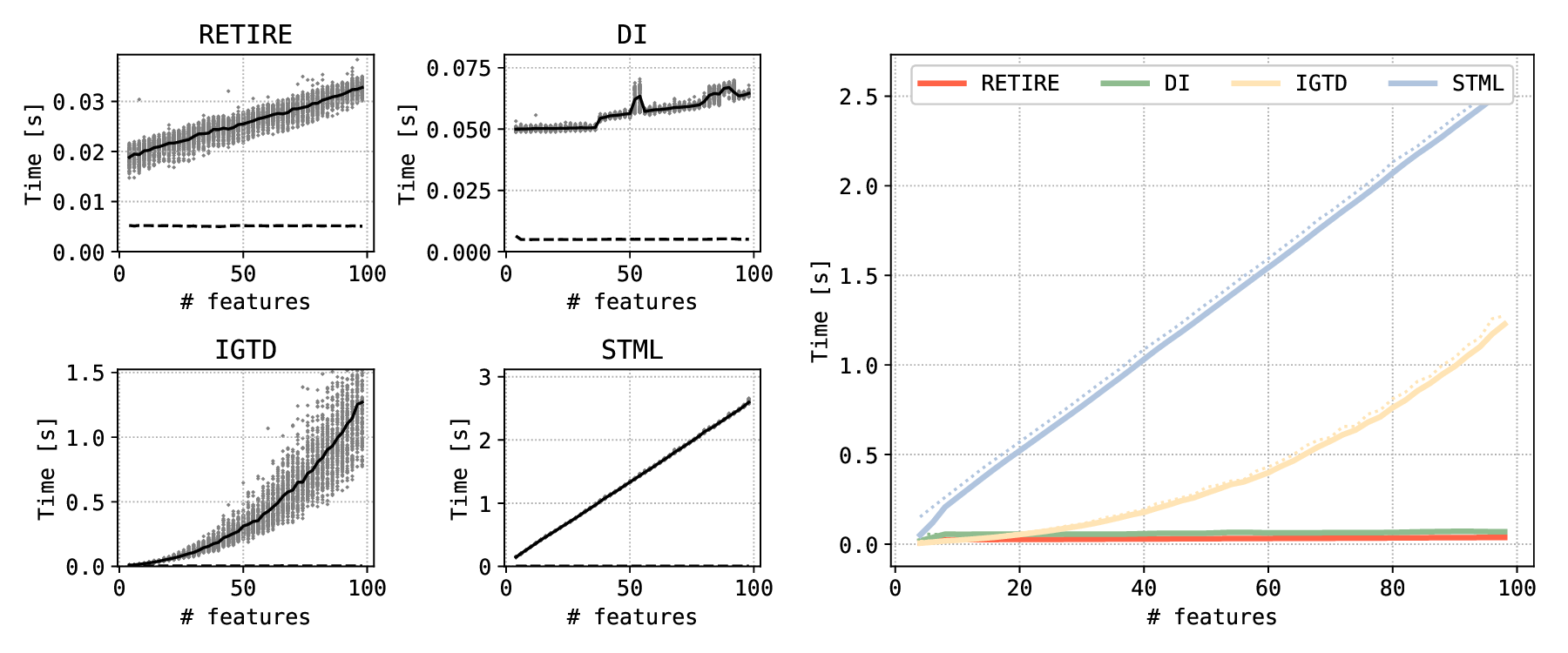}
    \caption{Time complexity analysis. Results for individual methods on the left and the comparison (with applied Simple Moving Average) on the right. 
       }
    \label{fig:timecomp}
\end{figure}

The computational complexity experiments were measured separately for data encoding time and model inference time. The results are shown in Figure \ref{fig:timecomp}. As can be seen, the lowest time values are achieved by the \textsc{retire} and \textsc{di} methods. Both graphs show a linear relationship between embedding preparation time and the number of features, although in the case of \textsc{di} -- in the first phases, it is flat. This can be justified by the fact that \textsc{di} performs optimization (\emph{Assymetric Greedy Search}), which can reach convergence in the first phases, but not for problems with higher dimensionality. As for \textsc{retire}, this linear time characteristic was an assumed theoretical, computational complexity that responds to \textbf{RQ4}. It can be explained that drawing the figure is related to an additional condition check with each new feature. Lastly, the \textsc{stml} also has linear computational complexity, although the time values achieved by the algorithm increase much faster.

An important observation is also the inference time, which, for most cases, is constant and very low when compared to the encoding time. This is an expected behavior since the structure of the \textsc{cnn} remains constant. The only differences are due to the processing time of smaller resolution images, which is relatable for \textsc{igtd}, although these are not noticeable differences. In addition, \textsc{igtd} is characterized by exponential computational complexity, which, with a large number of features, might not be feasible for computations.

\section{Conclusions}
The study aimed to expand the still under-exploited but extremely promising area of data transformation methods for tabular data analysis. This was done by proposing a novel \emph{multi-dimensional encoding} method, namely \emph{Radar-based Encoding from Tabular to Image REpresentation} (\textsc{retire}), which allows obtaining an image representation depicting the characteristics of the features of each problem instance as a single geometric shape.

Extensive comparison experiments have been carried out, showing that the representation obtained using \textsc{retire} in conjunction with the \emph{ResNet18} architecture enables a balanced accuracy score statistically significantly better than \emph{state-of-the-art} \textsc{mde} methods with publicly available implementations. Most importantly, \textsc{retire} is the only \emph{multi-dimensional encoding} method tested that is not statistically significantly inferior to the \emph{XGBoost} algorithm and even surpasses it in terms of average rank.

Additional strengths of the proposed method are its inherent explainability and relatively low linear computation time when compared to other \textsc{mde} approaches. Analysis using \textsc{shap} confirmed that the features considered most important by \emph{ResNet18} in the \textsc{retire} representation overlap with those that have the greatest impact on the classification results of the \emph{XGBoost} algorithm.

Future work may include a study of the transferability and explainability of \textsc{mde} methods to provide more insight into the specifics of their work, as well as the use of color in methods so far focused on binary or grayscale images.

\begin{credits}
\subsubsection{\ackname} This work was supported by the statutory funds of the Department of Systems and Computer Networks, Faculty of Information and Communication Technology, Wroclaw University of Science and Technology.

\subsubsection{\discintname} The authors have no competing interests to declare that are relevant to the content of this article.
\end{credits}
%
\bibliographystyle{splncs04}
\bibliography{bibliography}

\end{document}